\pdfoutput=1

\documentclass[11pt]{article}

\usepackage[final]{coling}

\usepackage{times}
\usepackage{latexsym}

\usepackage[T1]{fontenc}

\usepackage[utf8]{inputenc}

\usepackage{microtype}

\usepackage{inconsolata}

\usepackage{graphicx}
\usepackage{pifont}
\usepackage{ulem}
\usepackage{booktabs}
\usepackage{multirow}
\usepackage{tabularx}
\usepackage{hyperref}
\usepackage{multicol} 

\usepackage{amsmath}
\usepackage{xcolor}
\usepackage{listings}
\usepackage{colortbl}
\usepackage{tcolorbox}

\newcommand{\nestools}{\textsc{NesTools}~}

\title{\nestools: A Dataset for Evaluating Nested Tool Learning Abilities of Large Language Models}

\author{
Han Han, Tong Zhu, Xiang Zhang, Mengsong Wu, Hao Xiong, \textbf{Wenliang Chen}\thanks{\enspace Corresponding author}\\
~Institute of Artificial Intelligence, School of Computer Science and Technology, \\ Soochow University, China \\
\texttt{\{hhan,tzhu7,xzhangxzhang23,mswumsw,hxiongxionghao\}@stu.suda.edu.cn} \\
\texttt{wlchen@suda.edu.cn}
}

\begin{document}
\maketitle

\begin{abstract}

Large language models (LLMs) combined with tool learning have gained impressive results in real-world applications.
During tool learning, LLMs may call multiple tools in nested orders, where the latter tool call may take the former response as its input parameters.
However, current research on the nested tool learning capabilities is still under-explored, since the existing benchmarks lack relevant data instances.
To address this problem, we introduce \nestools to bridge the current gap in comprehensive nested tool learning evaluations.
\nestools comprises a novel automatic data generation method to construct large-scale nested tool calls with different nesting structures.
With manual review and refinement, the dataset is in high quality and closely aligned with real-world scenarios.
Therefore, \nestools can serve as a new benchmark to evaluate the nested tool learning abilities of LLMs.
We conduct extensive experiments on 22 LLMs, and provide in-depth analyses with \nestools, which shows that current LLMs still suffer from the complex nested tool learning task\footnote{Our code and dataset are available at \url{https://github.com/hhan1018/NesTools}}.

\end{abstract}
\section{Introduction}

\begin{figure}[t]
 \centering
  \includegraphics[width=0.9\linewidth]{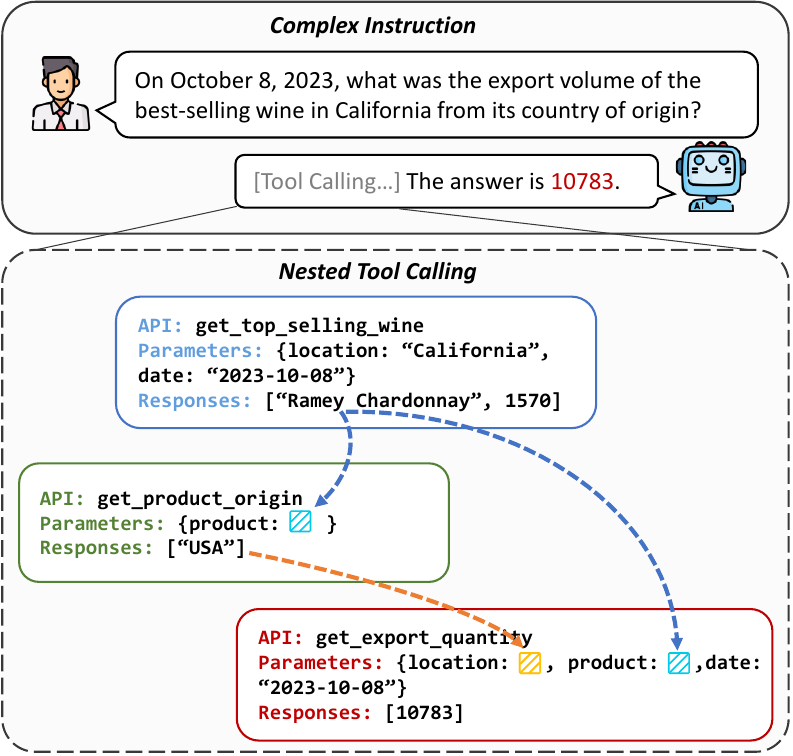}
  \caption {Example of nested tool calling.}
  \label{fig:nest-example}
\end{figure}

Large Language Models (LLMs) have shown powerful abilities in natural language understanding and reasoning \citep{achiam2023gpt, dubey2024llama, yang2024qwen2, zhu2024llama}.
To extend such abilities into real-world systems, tool learning~\citep{inaba2023multitool} has become a promising paradigm to solve complex problems and reduce hallucinations with external APIs, such as the calculator and the search engine \citep{patil2023gorilla,schick2024toolformer}.
In a real-world application, LLMs may interact with multiple tools~\citep{song2023restgpt, ye2024tooleyes, basu2024api, huang2023metatool}.
As shown in Figure~\ref{fig:nest-example}, the multi-tool calling process is nested, where the latter tool may take the former one's response as an input parameter.
This scenario is prevalent in accomplishing complex tasks with many execution steps.

However, the LLMs' nested tool learning ability is still under-explored, and corresponding benchmarks are absent to provide comprehensive evaluations.
Among current tool learning datasets, some datasets completely ignore the nested tool calls~\citep{tang2023toolalpaca, patil2023gorilla}.
Other benchmarks only have a small quantity of nested tool calls~\citep{li-etal-2023-api, huang2024planning, wu2024seal} or have low qualities with repetitive patterns~\citep{shen2023taskbench} and coarse evaluations~\citep{chen-etal-2024-eval,qin2023toolllm}. 
To this end, it is hard to comprehensively assess LLMs' performance on the real-world nested tool learning scenario, and provide insights for further model development.

\begin{table*}[t]
\centering
\renewcommand{\arraystretch}{1.0}
\resizebox{1.0\textwidth}{!}{
\begin{tabular}{clcccccc}
\toprule
\multicolumn{2}{c}{\textbf{}}                                  & \multicolumn{1}{c}{\begin{tabular}[c]{@{}c@{}}\textbf{\nestools}\\(Ours)\end{tabular}} & \multicolumn{1}{c}{\begin{tabular}[c]{@{}c@{}}\textbf{T-Eval}\\\citep{chen-etal-2024-eval}\end{tabular}} & \multicolumn{1}{c}{\begin{tabular}[c]{@{}c@{}}\textbf{API-Bank}\\\citep{li-etal-2023-api}\end{tabular}} & \multicolumn{1}{c}{\begin{tabular}[c]{@{}c@{}}\textbf{ToolBench}\\\citep{qin2023toolllm}\end{tabular}} & \multicolumn{1}{c}{\begin{tabular}[c]{@{}c@{}}\textbf{UltraTool}\\\citep{huang2024planning}\end{tabular}} & \multicolumn{1}{c}{\begin{tabular}[c]{@{}c@{}}\textbf{BFCL} \\ \citep{patil2023gorilla}
\end{tabular}}\\ \midrule
\multirow{2}{*}{\textbf{Tools}}   
                                    & \textbf{Amount}                 & 3,034                                    & 50                                          & 73                                  & 16,464       & 436   & 1,618                             \\
                                    & \textbf{Avg. params (required)} & 2.24                                  & 1.24                                            & 1.97                               & 1.01$^{\diamondsuit}$  &4.22  &2.11              \\ \midrule
\multirow{4}{*}{\textbf{Instances}} 
                                    & \textbf{Amount}                 & 1,000                                   & 553                                           & 485                                  & 126,486         & 1,000   & 2,000                         \\
                                    & \textbf{Multiple-tool callings}          & 1,000                                   & 553               &122                     & 85,330  $^{\diamondsuit}$                                 & 867    & 490                                \\
                                    & \textbf{Nested-tool callings}         & 830                                     & N/A$^{\spadesuit}$                                   & 50                             & N/A$^{\spadesuit}$                  & 227           & 0    \\
                                    & \textbf{Avg. tool call}         & 3.04                                     & 5.81                                   & 1.53                             & unknown                  & 2.38           & unknown
                                                                        \\\midrule
                                                                        
\multicolumn{2}{l}{\textbf{Multiple-tool scenario?}}                       & {\color[HTML]{008114}\ding{52}}                                     &  {\color[HTML]{008114}\ding{52}}                                             &    {\color[HTML]{008114}\ding{52}}                          &  {\color[HTML]{008114}\ding{52}}   & {\color[HTML]{008114}\ding{52}}     & {\color[HTML]{008114}\ding{52}}                             \\
\multicolumn{2}{l}{\textbf{Evaluation of nesting?}}          &  {\color[HTML]{008114}\ding{52}}                                     &{\color{red}\ding{56}}                                               & {\color{red}\ding{56}}                                   & {\color{red}\ding{56}}      &{\color{red}\ding{56}}        &{\color{red}\ding{56}}                               \\ \bottomrule
\end{tabular}
}
\caption{Comparison of tool learning datasets for evaluation.   $^{\spadesuit}$ The tool calling procedure is carried under a multi-step setting and the exact number cannot be obtained. $^{\diamondsuit}$ The statistics refer to Seal-Tools\citep{wu2024seal}.}
\label{dataset_comparsion}
\end{table*}

To address the above challenges, we introduce \nestools, a high-quality nested tool learning benchmark to provide comprehensive evaluations.
\nestools offers an innovative automated data construction scheme, generating large-scale and more diverse examples of nested tool learning compared to existing datasets.
The entire dataset construction process includes tool \& instance generation, query generation, and data review \& refinement.
To ensure an accurate and consistent evaluation, we carefully select and cross-verify 1,000 data entries.
For more comprehensive evaluations, we assess models on four deterministic dimensions: tool selection, tool calling order, parameter filling, and nested parameter filling.
Compared to existing benchmarks in Table~\ref{dataset_comparsion}, our dataset focuses on the nested tool learning task and provides large-scale tools with more nested calls.
In addition, our fine-grained assessment dimensions on nesting tool calls could provide a set of more comprehensive tool learning evaluations that are closely aligned with real-world scenarios.

We conduct extensive experiments on \nestools with 22 popular LLMs, including proprietary models and open-weight models.
We provide thorough analyses in terms of the nesting depth, nesting structure, scaled model sizes, and robustness effects.
The results show that although models benefit from size scaling, they still suffer from the simple tool selection, and the performance would degrade when tools are deeply nested.
Our core contributions are as follows:
\begin{itemize}
    \item We propose a novel automatic data construction pipeline to easily generate large-scale nested tool learning datasets.
    \item We introduce \nestools, a high-quality dataset with large-scale diverse examples for comprehensive nested tool learning evaluations.
    \item We conduct extensive experiments on 22 LLMs to verify their effectiveness and generalization abilities, providing detailed analyses and insights for LLMs in the field of nested tool learning.
\end{itemize}

\section{Related Work}

\paragraph{Tool learning}

Early works~\citep{yao2022react, schick2024toolformer, paranjape2023art} have incorporated straightforward tools such as search engines and calculators to enhance LLMs' access to up-to-date information and precise mathematical reasoning.
Following this, API-bank~\citep{li-etal-2023-api} constructs several tools and tool-use dialogues. ToolBench~\citep{qin2023toolllm} employs real-world APIs to construct datasets capable of addressing a broader spectrum of user queries. T-eval~\citep{chen-etal-2024-eval} collects tools from common domains, leveraging a multi-agent paradigm to resolve solution annotations. UltraTool~\citep{huang2024planning} starts with real-world queries that may require the construction of new tools.

\begin{figure*}[t]
 \centering
  \includegraphics[width=\linewidth]{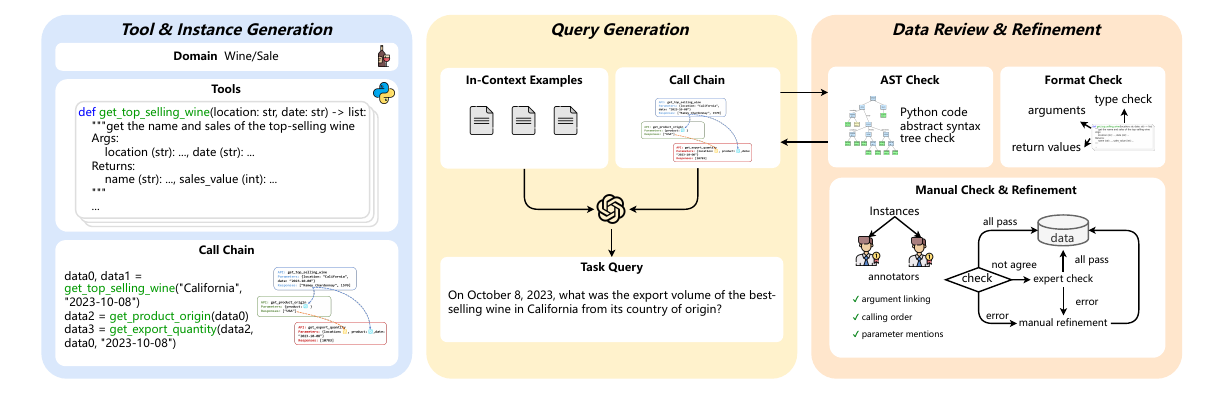}
  \caption {The dataset construction process of \nestools.}
  \label{fig:liuchengtu}
\end{figure*}

\paragraph{Tool Evaluation}
Conducting reasonable and effective evaluations of tool learning capabilities for Large Language Models (LLMs) is essential~\citep{qin2023tool, qu2024tool}. API-Bank~\citep{li-etal-2023-api} evaluates from perspectives of the correctness of API calls and the quality of LLMs' responses. ToolBench~\citep{qin2023toolllm} uses coarse-grained evaluation metrics including pass rate and win rate backed up by ChatGPT. UltraTool~\citep{huang2024planning} evaluates from perspectives of planning, tool creation, and tool usage. Although these benchmarks include a limited number of nested tool calling, researchers have not undertaken a comprehensive analysis and evaluation of this phenomenon.
Consequently, the performance of LLMs in handling nested tool calling remains under-explored.

To address this gap, we introduce \nestools, a dataset specifically constructed to feature a multitude of nested tool calling scenarios across various domains. The purpose of this dataset is to systematically evaluate the performance of LLMs in nested tool learning abilities.

\section{Dataset Construction}

Leveraging large language models (LLMs) for data generation has become quite common in various domains~\citep{wang2023self,Yu2024ReliableDG}. It is very necessary to explore a feasible and comprehensive data generation scheme for nested tool calling.

To obtain \nestools, we propose an automatic dataset generation pipeline with manual reviews.
As shown in \figurename~\ref{fig:liuchengtu}, the whole process starts with 
tool and instance generation ($\S \ref{sec:Tool/Instance Generation}$), followed by query generation ($\S \ref{sec:Query Generation}$), and data review \& refinement ($\S \ref{sec:manual review and refinement}$).
Lastly, we provide a comparative analysis for \nestools with other datasets ($\S \ref{sec:Dataset Summary}$).

\subsection{Tool \& Instance Generation}
\label{sec:Tool/Instance Generation}

In real-world applications with a limited number of tools, call chains containing nested calls are relatively restricted and homogenized, and there are not too many non-duplicated and diverse call chains. To get a wide range of tools and instances, existing benchmarks mostly use a conventional two-round generation scheme, including filtering or constructing a portion of the tools first, and then filtering or generating some instances~\citep{qu2024tool,chen-etal-2024-eval}.
However, the proportion of generated samples containing nested instances under this scheme is very small because LLMs cannot synthesize as many candidate tools as possible to obtain potential nested call chains under the limitations of the context.
If relying on manual brainstorming, this is very costly and difficult to scale and ensure diversification.

How to increase the frequency of nested calling instances? LLMs are relatively unfamiliar with nested tool calling. However, they are very familiar with nested functions in code because the pre-training process has already injected large amounts of code knowledge into LLMs~\citep{dubey2024llama}.
The two types of nesting can be very similar to some extent, with the common code pattern being to define a number of Python functions and then execute the calls in the main function.
In this task, we ask GPT-3.5 to generate fixed-format functions as tools with annotation for necessary tool information, and then generate a call chain to execute the tools in the main function.

To ensure that the automated generation tools can cover a wider range of domains and minimize repetition rates as much as possible, we collect the domains from Seal-Tools~\citep{wu2024seal}, including 146 domains and 5,860 sub-domains, serving as a reference guide for the generation of the subsequent tools and instance.

When generating tools, we adopt a template with functions in Python code, including comments that describe the tool, required parameters, optional parameters, and return parameters. Initially, we attempt a nested call pattern using ``func1(func2(...), *args)''. However, this style is insufficient for more complex nested examples, as it results in lengthy expressions and lacks clarity in referencing specific return parameters when dealing with multiple return values. The improved generation pattern is as follows: ``data0, ... = func1(...), data2, ... = func2(data0, ...)''. This approach accommodates more complex nesting and indicates specific return values, as illustrated in Figure~\ref{fig:liuchengtu}.

It is worth mentioning that the in-context examples are very important, there are different nested structures for different nested calls, to ensure the diversity and rationality of the generation, we carefully select a portion of high-quality samples to form the sample pool, and rotate these examples during the iterative generation process.
At the same time, to minimize homogeneous generation, we set the generation temperature to 0.95 to seek for more diversified generation. The prompt and example can be found in Appendix~\ref{Dataset Construction details: Tool/Instance Generation}.

\subsection{Query Generation}
\label{sec:Query Generation}

While the tool and call chain are generated, we reserve a comment to initially generate some raw queries that refer to the logic and flow of the call chain, although they may have flaws such as missing necessary parameter values.
We then provide the entire actual call chain and the raw query to allow GPT-3.5 to generate the complete task, a step that allows the big model to populate all necessary parameters. Finally, we let GPT-4~\citep{achiam2023gpt} optimize and rewrite the query to make it more complete and closer to the real requirements. 
Corresponding prompts can be found in Appendix~\ref{Dataset Construction details: Query Normalization} and Appendix~\ref{Dataset Construction details: Query Refinement}.

\subsection{Data Review \& Refinement}
\label{sec:manual review and refinement}

To ensure the data quality, we perform strict reviewing steps and make further refinements.
We first perform automatic checks. Given that nested tool calling contains multiple tools where each tool contains multiple return values, it is necessary to ensure that each return value in the call chain has a variable to receive it.
As a result, we conduct the format check to filter out abnormal output formats and irregularities in generation.
It is worth noting that in addition to the format error, the automated screening includes an abstract syntax tree (AST) check to ensure the consistency of tools and call chains.
This process could align the return values and parameters between the candidate tools and the call chains.

After the automated check, we perform a manual review for further quality control.
For the call chains, we mainly check whether the values of the parameters are in accordance with the description information in the golden tools.
Besides, the most critical aspect of this reviewing step is to determine whether the nested parameters are filled in correctly.
It requires manual tracing back to the parent node tool that returns the parameter, and carefully comparing the parent node's return value with the textual description of the parameter at this position.
Potential errors include the parameter being a non-nested parameter but being filled as a nested one, or being a nested parameter but being filled with an incorrect return value from a previous tool.

For the task query review, we mainly check whether the non-nested parameters in the call chain are mentioned in the query, whether the query description conforms to the logic of the call chain, and whether the low-quality samples are optimized to meet real-world requirements.
We employ two annotators to tag \nestools from two perspectives: consistency between the call chain and the tools, and whether the query meets the requirements.
Besides, the annotators are required to refine the data instances if there are errors.
If there is a disagreement between the two annotators, an expert would help make a final annotation.
The overall annotation process shows a high agreement of 0.96. More details can be found in Appendix~\ref{Dataset Construction details: Other details}.

\subsection{Dataset Summary}
\label{sec:Dataset Summary}

As shown in \tablename~\ref{dataset_comparsion}, we compare \nestools in detail with the currently available datasets, and with the necessary number of tools and instances guaranteed, \nestools has a clear advantage in terms of nested tool calling, which not only covers a higher percentage of nesting and includes a new framework of evaluation LLMs for nested tool calling.

\section{Experiments}

\begin{table*}[t]
\centering
\begin{tabularx}{\textwidth}{l *{13}{>{\centering\arraybackslash}X}}
\toprule
\multicolumn{1}{l}{\multirow{2}{*}{\textbf{Model}}} & \multicolumn{3}{c}{\textbf{Selection}} & \multicolumn{3}{c}{\textbf{Order}} & \multicolumn{3}{c}{\textbf{Parameter}} & \multicolumn{3}{c}{\textbf{Nested Param}} &\multirow{2}{*}{\textbf{Avg.}} \\
\cmidrule(lr){2-4} \cmidrule(lr){5-7} \cmidrule(lr){8-10} \cmidrule(lr){11-13} 
 & P & R & F1 & P & R & F1 & P & R & F1 & P & R & F1 \\
\midrule
\rowcolor{red!20} \multicolumn{14}{c}{\textit{Proprietary Models}} \\

GPT-4o      & \textbf{85.1} & \textbf{86.2} & \textbf{85.7} & \textbf{72.9} & \textbf{74.3} & \textbf{73.6} & \textbf{75.5} & \textbf{78.6} & \textbf{77.0} & \textbf{73.5} & \textbf{69.3} & \textbf{71.3} & \textbf{76.9}\\
GPT-4o-mini & 77.3 & 72.6 & 74.9 & 60.6 & 56.8 & 58.6 & 65.3 & 64.6 & 64.9 & 58.8 & 48.1 & 52.9 & 62.8 \\
GPT-3.5     & 69.0 & 71.2 & 70.1 & 47.9 & 50.4 & 49.1 & 56.7 & 62.0 & 59.2 & 46.5 & 41.7 & 44.0 & 55.6 \\
Claude-3.5  & 79.6 & 82.8 & 81.2 & 64.3 & 68.2 & 66.2 & 70.6 & 75.9 & 73.2 & 68.3 & 65.5 & 66.8 & 71.8 \\

\midrule
\rowcolor{green!20} \multicolumn{14}{c}{\textit{Open-Weight Models}} \\

LLaMA3.1-8B   & 65.8 & 71.7 & 68.6 & 44.3 & 50.7 & 47.3 & 44.3 & 51.1 & 47.5 & 40.8 & 42.7 & 41.7 & 51.3 \\
LLaMA3.1-70B  & 82.3 & 81.9 & 82.1 & 68.9 & 68.8 & 68.8 & 57.7 & 60.3 & 59.0 & 67.6 & 61.1 & 64.1 & 68.5 \\
LLaMA3.1-405B & \textbf{87.9} & \textbf{87.2} & \textbf{87.5} & \textbf{77.3} & \textbf{77.4} & \textbf{77.3} & 66.4 & 68.5 & 67.4 & \textbf{78.8} & \textbf{72.7} & \textbf{75.7} & \textbf{77.0} \\
Mistral-7B v0.2    & 49.6 & 44.1 & 46.7 & 22.0 & 21.8 & 21.9 & 35.2 & 33.0 & 34.1 & 21.9 & 14.1 & 17.1 & 29.9 \\
Mixtral-8x7B  & 62.4 & 64.3 & 63.4 & 38.0 & 41.3 & 39.6 & 46.3 & 49.9 & 48.0 & 35.6 & 35.1 & 35.3 & 46.6 \\
Mixtral-8x22B & 70.4 & 75.1 & 72.7 & 49.6 & 54.7 & 52.0 & 59.4 & 66.8 & 62.9 & 53.0 & 52.3 & 52.6 & 60.1 \\
Qwen2-0.5B    & 0.0  & 0.0  & 0.0  & 0.0  & 0.0  & 0.0  & 0.0  & 0.0  & 0.0  & 0.0  & 0.0  & 0.0  & 0.0  \\
Qwen2-1.5B  & 39.6 & 18.1 & 24.8 & 12.4 & 4.4  & 6.5  & 28.4 & 14.2 & 18.9 & 0.0  & 0.0  & 0.0  & 12.5 \\
Qwen2-7B    & 59.4 & 58.9 & 59.1 & 34.9 & 35.5 & 35.2 & 41.1 & 43.1 & 42.1 & 29.5 & 14.1 & 19.1 & 38.9 \\
Qwen2-57B     & 63.8 & 68.9 & 66.2 & 40.2 & 45.5 & 42.7 & 47.6 & 54.3 & 50.7 & 37.8 & 29.0 & 32.8 & 48.1 \\
Qwen2-72B     & 78.4 & 79.2 & 78.8 & 62.9 & 64.2 & 63.6 & \textbf{67.0} & \textbf{69.9} & \textbf{68.4} & 57.9 & 58.6 & 58.3 & 67.3 \\
Qwen1.5-0.5B  & 10.2 & 0.2  & 0.3  & 0.0  & 0.0  & 0.0  & 2.2  & 0.1  & 0.1  & 0.0  & 0.0  & 0.0  & 0.1  \\
Qwen1.5-1.8B    & 21.2 & 3.4  & 5.8  & 2.5  & 0.4  & 0.8  & 9.8  & 1.7  & 2.9  & 0.0  & 0.0  & 0.0  & 2.4  \\
Qwen1.5-7B    & 55.0 & 35.5 & 43.2 & 27.8 & 18.7 & 22.3 & 38.0 & 27.1 & 31.7 & 21.2 & 5.2  & 8.4  & 26.4 \\
Qwen1.5-14B   & 61.8 & 56.9 & 59.3 & 37.0 & 35.3 & 36.1 & 46.4 & 45.8 & 46.1 & 34.2 & 21.4 & 26.3 & 41.9 \\
Qwen1.5-32B   & 65.2 & 61.1 & 63.1 & 41.7 & 40.7 & 41.2 & 52.3 & 51.3 & 51.8 & 45.6 & 27.2 & 34.1 & 47.5 \\
Qwen1.5-72B   & 66.0 & 59.7 & 62.7 & 43.1 & 40.0 & 41.5 & 49.4 & 47.4 & 48.4 & 31.6 & 21.5 & 25.6 & 44.6 \\
Qwen1.5-110B  & 69.7 & 65.0 & 67.3 & 49.0 & 46.7 & 47.8 & 56.3 & 54.8 & 55.6 & 44.8 & 31.8 & 37.2 & 52.0 \\

\bottomrule
\end{tabularx}
\caption{The main results of \nestools. Avg. is the average F1 score of all evaluation metrics. \textbf{Bold} represents the best score among all models.}
\label{table: main results}
\end{table*}

\subsection{Experimental Setup}

In order to evaluate the LLM's understanding of the whole call chain of nested tool calls, 
we use the pattern of prompting for evaluation. 
For the parameter of the nested tools, the model is constrained to generate a placeholder like "API\_call\_\{number\}" to fill the parameter of nested position in the evaluation prompt.

For each sample, in order to simulate the scenario when the LLM faces similar tools, we utilize the gte-large retriever \footnote{\url{https://huggingface.co/Alibaba-NLP/gte-large-en-v1.5}} to retrieve 5 similar tools for each gold tool from the tool pool. To minimize the occurrence of ambiguous tools, we control for the fact that the name of each retrieved tool must meet a certain Levenshtein distance requirement with the names of all gold tools.

We conduct experiments on \nestools with the latest proprietary and open-weight LLMs. For the proprietary LLMs, we selected four representative LLMs: GPT-4o, GPT-4o-mini and GPT-3.5 from OpenAI and Claude3.5 from Anthropic\footnote{The version for GPT-4o is gpt-4o-2024-08-06, for GPT-4o -mini is gpt-4o-mini-2024-07-18, for GPT-3.5 is gpt-3.5-turbo-0125 and for Claude3.5 is claude-3-5-sonnet-20240620.}. For the open-weight LLMs, we choose the representative ones: LLaMA3.1~\citep{dubey2024llama}, Qwen1.5~\citep{bai2023qwen}, Qwen2~\citep{yang2024qwen2}, Mistral~\citep{jiang2023mistral} and Mixtral~\citep{jiang2024mixtral}. We perform detailed experiments on each of their sizes.

The detailed evaluation prompt is shown in Appendix~\ref{Evaluation details: Evaluation Prompt}.

\subsection{Evaluation Metrics}

For \nestools, to standardize the evaluation, we devise metrics for the following aspects: correctness of tool selection, correctness of the order of tool calls, correctness of parameter filling, and correctness of nested parameter filling.
Each evaluation is a deterministic P/R/F1 metric. 

\textbf{Selection P/R/F1} measures the accuracy of LLM's tool selection, which is a common evaluation dimension.

\textbf{Order P/R/F1} measures the accuracy of LLM's judgment of the previous and subsequent tools. Since nested tool calling includes the use of the return value of the previous tool, it is necessary to evaluate the order of the tools before and after.

\textbf{Parameter P/R/F1} measures the correctness of LLM's filling parameters, which is also a common evaluation dimension.

\textbf{Nested Param P/R/F1} is specific to nested calls and evaluates the accuracy of parameters at nested locations in the call chain.

Other metrics and details are listed in Appendix~\ref{Evaluation details: Other Evaluation Metrics}.

\subsection{Main Results}

In the field of tool learning, for a long time, the GPT-4 series has occupied the position of the best performance, and open-weight LLMs are once out of reach \citep{huang2024planning,chen-etal-2024-eval}. However, as shown in Table~\ref{table: main results}, we are pleasantly surprised to find that LLaMA3.1-405B slightly outperforms GPT-4o on \nestools, achieving the best performance among all evaluated LLMs. This signifies that open-weight and proprietary LLMs are currently very competitive in tool learning. 

We observe that the LLaMA3.1 series exhibits significant advantages in Selection and Order. Notably, LLaMA3.1-8B outperforms LLMs like Mixtral-8x7B and Qwen2-57B, which are substantially larger in size. Qwen2-72B reaches the best performance in parameters among open-weight LLMs. Mixtral-8x22B has the highest format accuracy among open-weight LLMs, second only to GPT-4o and Claude-3.5 in all LLMs. 

However, in terms of parameter filling, the LLaMA3.1 series shows an unusual disadvantage. Analysis of sample cases reveals that LLaMA3.1 is less sensitive to parameter types, often generating incorrect parameter types in JSON outputs, even though the extracted answer information may be correct. This issue is most pronounced in the LLaMA3.1 series. LLaMA3.1-70B is most severely affected, resulting in a loss of nearly 10 percentage points in parameter metric. It is worth noting that such problems are significantly less prevalent in other LLMs, and they are entirely absent in Claude-3.5. This suggests that compared to other LLMs, the LLaMA series may have undergone relatively less training focused on JSON format alignment.

Nested tool calling presents a certain level of difficulty, and the stronger the LLM's capabilities, the better it performs in resolving nested tool calling. Filling nested parameters requires the model to truly understand the call chain and identify which specific return value corresponds to each nested parameter. Judging from the metrics of parameters and nesting, LLM's performance in nesting is generally lower than that in parameters, but this gap is narrowing as the LLM's overall performance improves.

\section{Further Analysis}

\subsection{Scaling Analysis}

\begin{figure}[t]
 \centering
  \includegraphics[width=\linewidth]{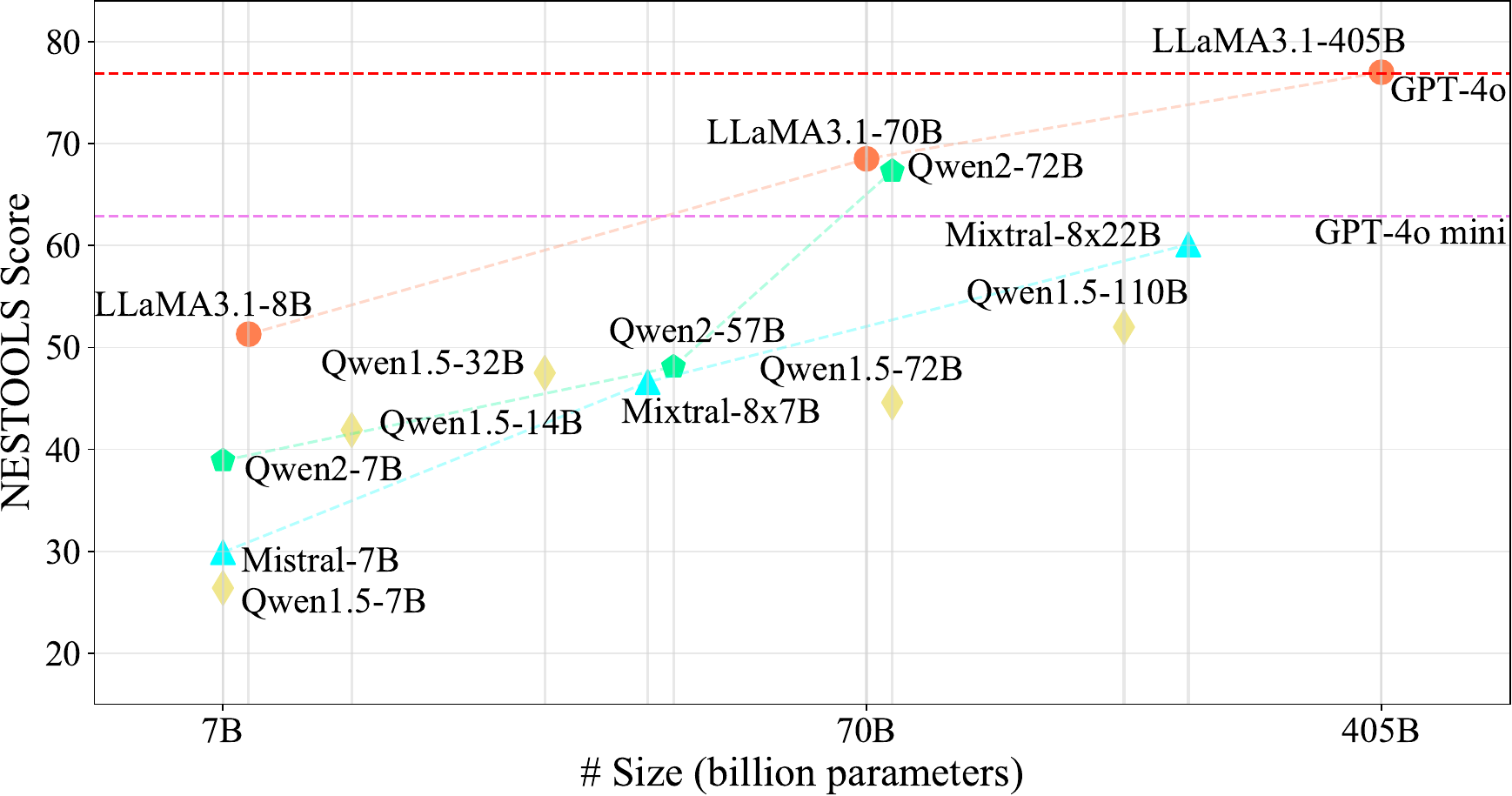}
  \caption{Model scaling results on \nestools.}
  \label{fig:scaling comparison}
\end{figure}

Scaling law is common in LLMs. As shown in Figure~\ref{fig:scaling comparison}, we can conclude that most LLMs such as Qwen2, LLaMA3.1, Mistral, and GPT-4o series follow the scaling law in \nestools. The larger the size, the stronger its performance in tool calling, which is consistent with the previous research conclusions. However, it is worth mentioning that the Qwen1.5 series does not fully follow the scaling law. Compared with the initially released Qwen1.5 series, Qwen team released Qwen1.5-32B two months later. We speculate that this process may include more high-quality data and more optimized iterative training.

\begin{table*}[t]
\centering
\begin{tabularx}{\textwidth}{l *{13}{>{\centering\arraybackslash}X}}
\toprule
\multicolumn{1}{l}{\multirow{2}{*}{\textbf{Setting}}} & \multicolumn{3}{c}{\textbf{Selection}} & \multicolumn{3}{c}{\textbf{Order}} & \multicolumn{3}{c}{\textbf{Parameter}} & \multicolumn{3}{c}{\textbf{Nested Param}} &\multirow{2}{*}{\textbf{Avg.}} \\
\cmidrule(lr){2-4} \cmidrule(lr){5-7} \cmidrule(lr){8-10} \cmidrule(lr){11-13} 
 & P & R & F1 & P & R & F1 & P & R & F1 & P & R & F1 \\
\midrule

base & \textbf{65.8} & 71.7 & \textbf{68.6} & \textbf{44.3} & 50.7 & \textbf{47.3} & \textbf{44.3} & 51.1 & \textbf{47.5} & \textbf{40.8} & 42.7 & \textbf{41.7} & \textbf{51.3} \\
w/o type  & 62.7 & \textbf{73.6} & 67.7 & 40.5 & \textbf{51.4} & 45.3 & 42.4 & 52.7 & 47.0 & 35.7 & 43.1 & 39.0 & 49.8 \\
w/o conjunction & 59.8 & 73.4 & 65.9 & 35.7 & 48.3 & 41.0 & 41.1 & \textbf{53.4} & 46.5 & 33.2 & \textbf{43.8} & 37.7 & 47.8 \\

\bottomrule
\end{tabularx}
\caption{Robustness analysis conducted on LLaMA3.1-8B, base represents the results of the original setting, w/o type represents masking the type of response parameters of each tool, and w/o conjunction represents removing the conjunctions of the query.}
\label{tab:Robustness analysis}
\end{table*}

\subsection{Nesting Depth Analysis}

Different call chains have different call depths. We consider the entire call chain's nesting depth to be 1 if the instance contains no nested calls. For nested tool calling instances, each tool is at a specific nesting depth. Tools that do not use the return value of previous tools are at depth 1. If a tool uses the return value of a previous tool, its depth is the previous tool's depth plus one, and so on. For an instance, its overall nesting depth matches the deepest level among the tools.

\begin{figure}[t]
 \centering
  \includegraphics[width=\linewidth]{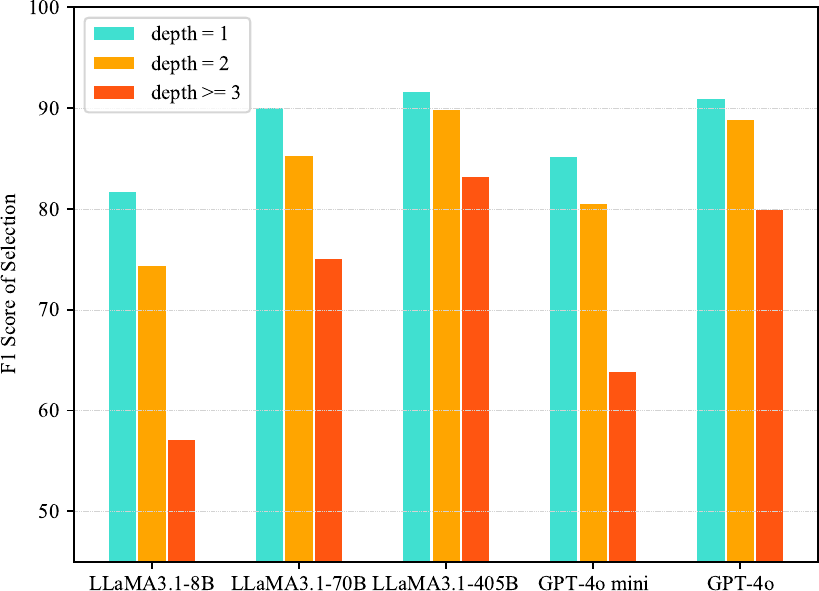}
  \caption {The relation between nesting depth and Selection F1 among LLMs.}
  \label{fig:depth analysis}
\end{figure}

We analyze the performance of LLMs on \nestools with different nesting depths. As shown in Figure~\ref{fig:depth analysis}, although the performance of the different LLMs varies, each of them shows a tendency that the deeper the nesting depth, the worse the performance of LLMs. This undoubtedly illustrates the difficulty of nested tool calling and its importance for multiple tool calling.

This trend can be attributed to several factors. Firstly, as the nesting depth increases, the complexity of the task grows exponentially. Each additional layer of nesting introduces new dependencies and potential points of failure, making it more challenging for the LLM to maintain coherence and accuracy throughout the entire call chain. Secondly, deeper nesting requires the model to retain and process more contextual information over longer sequences, which can lead to issues with memory and attention span. This is particularly relevant for LLMs that have fixed-length context windows, where the ability to handle long-range dependencies diminishes as the sequence length increases.

\subsection{Nesting Structure Analysis}

\begin{figure}[t]
 \centering
  \includegraphics[width=\linewidth]{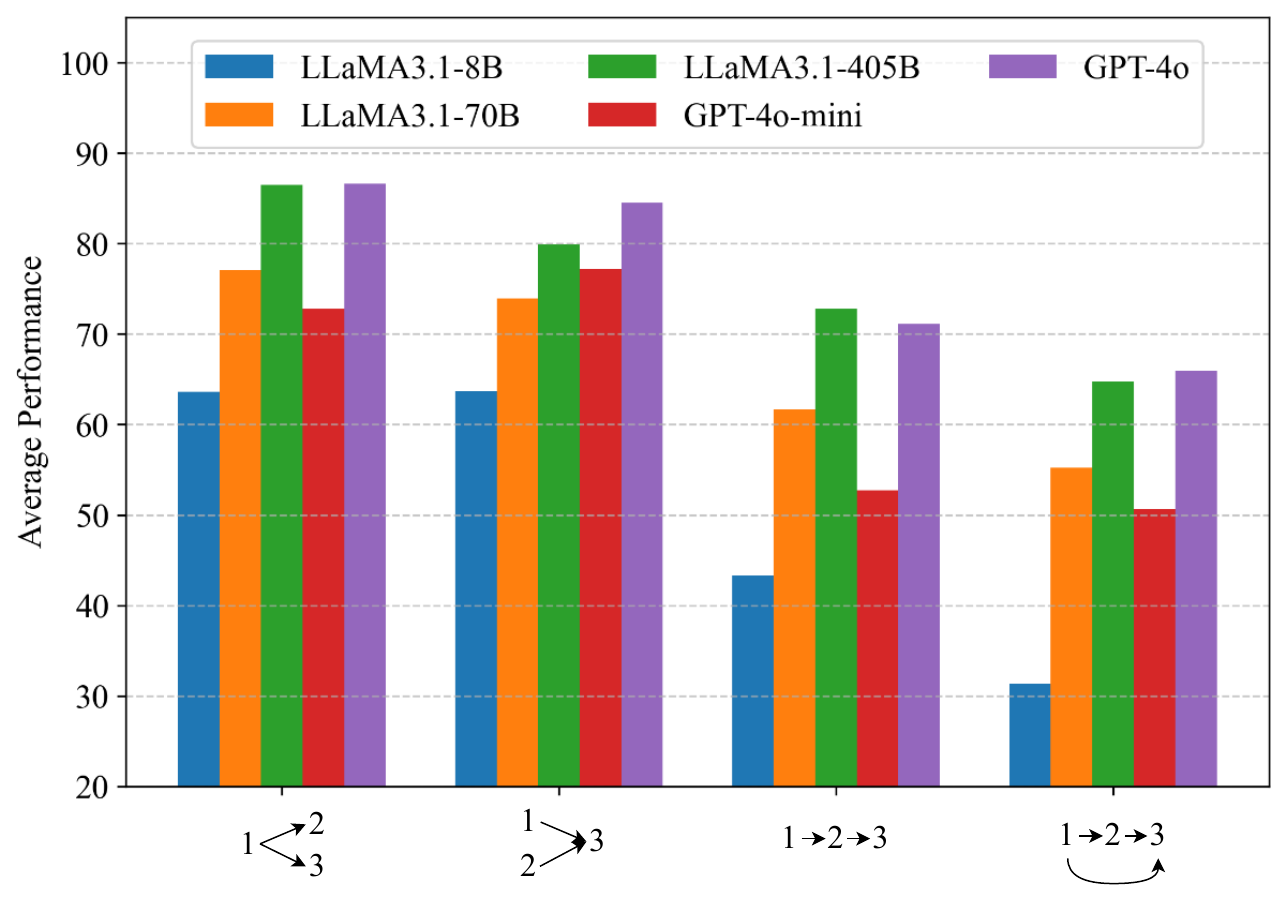}
  \caption {The averaged performance of different nesting structures. The arrow between two numbers indicates the nesting shape. For example, the first structure type (1$\rightarrow$2,1$\rightarrow$3) denotes that the 1st tool call's response contributes the input parameters for both the 2nd and the 3rd tool calls.}
  \label{fig:nesting-structure}
\end{figure}

As shown in Figure~\ref{fig:nesting-structure}, we observe a similar phenomenon with the nesting depth experiment, where shallow tool calls result in higher average performance.
However, the averaged performance would be higher if there is a leading tool call (1$\rightarrow$2,1$\rightarrow$3) that is followed by two separate calls (1$\rightarrow$3,2$\rightarrow$3).
Besides the long nesting call structure (1$\rightarrow$2$\rightarrow$3), the additional nesting connection ($1\rightarrow3$) at the most right column obtain the lowest performance among all the structures, which shows the adverse effect of structure complexity.

\subsection{Robustness Analysis}
\label{sec:Robustness analysis}

\begin{figure*}[t]
 \centering
  \includegraphics[width=\linewidth]{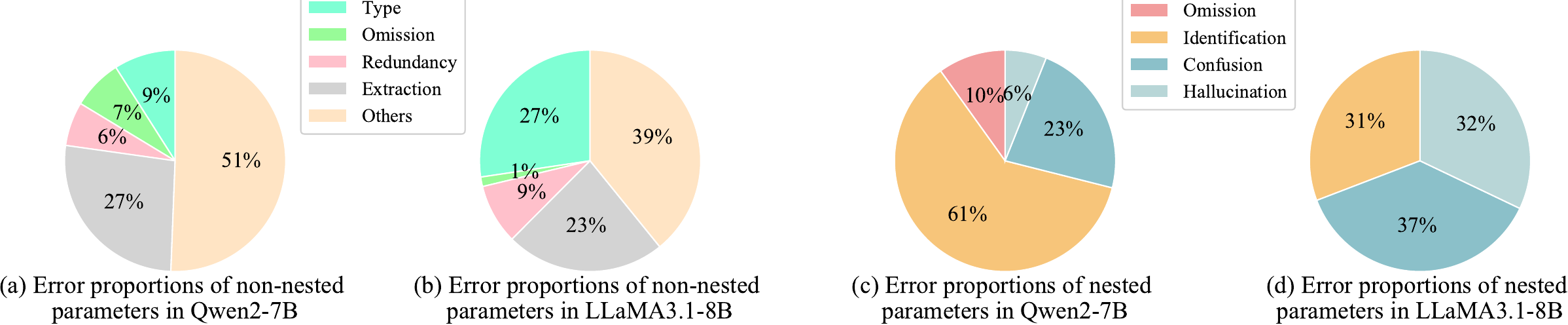}
  \caption {Parameter error analysis for Qwen2-7B and LLaMA3.1-8B}
  \label{fig:para error}
\end{figure*}

Although existing benchmarks have comprehensively evaluated the tool calling ability of LLMs, there is a lack of robustness analysis for tool calling based on these benchmarks. We intend to analyze the robustness of tool calls from both tool and query perspectives on \nestools.

From the query perspective, we notice that some conjunctions that do not affect the meaning often appear in the query of tool calls, such as "Then, After that, Please", etc. We guess whether the model works well because of the existence of a shortcut. Considering that the rule-based screening cannot completely cover the connectives in the query, we intend to remove these additional words and restore an original direct query. Considering that the rule-based screening cannot completely cover the connectives in the query, we intend to ask LLMs to remove these additional words. The prompt can be found in Appendix~\ref{Evaluation details: Prompt of removing conjunctions}. To ensure that this operation does not affect the core information of the query, we check some samples and don't find biases caused by this process.

From the tool perspective, we would like to explore the impact of tool integrity on the experimental results when the tool is missing a small part of the information, in order to keep consistency with the original evaluation process, and to ensure that the core information of the tool is complete, we only mask out the type of the return value of the tool to explore the impact of tool completeness on the experimental results.

As shown in Table~\ref{tab:Robustness analysis}, In the case of masking off the type of the tool return value, LLaMA3.1-8B shows a slight decrease in all metrics. However, In the case of removing the connectives in the query, the performance drops more significantly, especially the Order metric, which proves that the conjunction does help LLMs understand the order of the call chain. An interesting observation is that after applying additional factors, the R-value metrics generally increase, indicating that LLM tends to call more tools to cover the gold standard answers. In summary, it can be inferred that the robustness of current LLMs in the field of tool calling is still insufficient and worth exploring.

\subsection{Error Analysis}

We analyze the errors from LLMs during the evaluation process, unfolding both the tool and parameter perspectives.

\subsubsection{Tool Error}
We summarize the errors in the tool perspective and divide them into five categories: tool selection, tool omission, tool redundancy, hallucinations, and generation format.
Since these types of errors are usually mixed, it is hard to provide quantitative analysis.
To this end, we provide each error category with a case study in Appendix~\ref{sec:error-examples}.

Here are our overall findings:
\textit{(1)} \textbf{Tool selection}:
We find that similar tools may confuse the model, leading to errors in tool selection.
\textit{(2)} \textbf{Tool Omission}:
Although the golden tools corresponding to each query are included in the evaluation prompt, the model may still miss some key tools.
\textit{(3)} \textbf{Tool Redundancy}:
LLMs sometimes call redundant tools that are not task-related.
\textit{(4)} \textbf{Hallucinations}:
When faced with some specific tasks, the model may hallucinate and construct tools on its own that do not exist in the tool pool.
\textit{(5)} \textbf{Generation Format}:
During the evaluation process, some LLMs may generate results in the wrong format, which cannot be parsed for evaluations. The results regarding the accuracy of the format, as presented in Appendix~\ref{Evaluation details: Other Evaluation Metrics}, reveal that smaller models are more inclined to generate content in incorrect formats.

\subsubsection{Parameter error}

For errors in the parameter section, we analyze the error types from two perspectives: \textit{(1)}\textbf{Non-nested parameter} errors include parameter omissions (Omission), redundant parameters (Redundancy), the parameter value is correct but the type is incorrect (Type), the parameter value is wrongly extracted (Extraction), and other errors (Others). 
\textit{(2)}\textbf{Nested parameter} errors include non-nested parameter mistaken for nested (Hallucination), nested parameter not involved (Omission), nested parameter recognized as non-nested parameter (Identification), nested parameter identified but take an incorrect value (Confusion).

As shown in Figure~\ref{fig:para error}, we count the proportion of different error categories of Qwen2-7B and LLaMA3.1-8B.
For non-nested parameters, LLaMA3.1-8B tends to mispredict parameter types, while Qwen2-7B makes fewer mistakes.
For nested parameters, the most common mistake made by Qwen2-7B is recognizing nested parameters as non-nested parameters.
The errors made by LLaMA3.1-8B are relatively average, and they are more likely to fill in the wrong nested parameter values after identifying them.
This shows that LLaMA3.1-8B is more capable of identifying nested tool calling than Qwen2-7B to some extent.

\section{Conclusion}

In this paper, we introduce \nestools, a large-scale dataset of high-quality for evaluating the LLMs' nested tool learning abilities.
We propose a novel automatic method to generate cases involving nested tool calls. Through a multi-step process of manual review and refinement, we ensure the quality of \nestools. Statistical analysis reveals that our dataset spans a wide range of domains. Furthermore, experiments conducted on 22 LLMs demonstrate the significance of this task. Our findings indicate that LLMs still struggle with effectively handling nested tool calls.
We hope that \nestools, combined with our detailed experiments, can provide valuable insights and inspire future in-depth exploration of practical applications in tool agents.

\section*{Limitations}

This study mainly builds complex nested tool calling instances, providing a new perspective on nested tool calling. Although promising, in reality, there may be situations where the necessary preceding tools are missing, and the model needs to use its own capabilities to perform additional processing on the return value. Another limitation is that the tools involved in this study cannot be executed. Although these tools are not specific implementations, they can be used as simulation representations to accurately describe the functions of the tools, which can be used to evaluate the capabilities of LLMs in tool learning and provide guidance for the future development of actual tools. Later, we will try to design real execution interfaces for these tools to better simulate nested tool calls in real situations.
\section*{Acknowledgments}

This work is supported by the National Natural Science Foundation of China (Grant No. 62376177, 62261160648) and Provincial Key Laboratory for Computer Information Processing Technology, Soochow University. This work is also supported by Collaborative Innovation Center of Novel Software Technology and Industrialization, Project Funded by the Priority Academic Program Development of Jiangsu Higher Education Institutions. We would also like to thank the anonymous reviewers for their insightful and valuable comments.

\bibliography{custom}

\clearpage
\appendix

\section{Dataset Construction details}

\subsection{Tool/Instance Generation}
\label{Dataset Construction details: Tool/Instance Generation}

The corresponding prompt is presented in Table~\ref{tabs:Tool/Instance Generation}. An example for the generation process can be found in Table~\ref{tabs:ICL example for Tool/Instance Generation}.

\subsection{Query Normalization}
\label{Dataset Construction details: Query Normalization}

The corresponding prompt is presented in Table~\ref{tabs: Query Normalization}.

\subsection{Query Refinement}
\label{Dataset Construction details: Query Refinement}

The corresponding prompt is presented in Table~\ref{tabs: Query Refinement}.

\subsection{Other details}
\label{Dataset Construction details: Other details}

Based on predefined review criteria, the final agreement is computed by averaging the concordance between query and call chain verification. $N_T$ means that the data is considered error-free by all annotators, and data with $N_F$ is considered to have errors by all annotators. \( N_{overall} \) represents the total number of data. The final agreement is defined as:
\[ p = \frac{N_T + N_F}{N_{overall}} \]

\section{Evaluation details}

\subsection{Other Evaluation Metrics}
\label{Evaluation details: Other Evaluation Metrics}

In addition to the four metrics of \textbf{Selection}, \textbf{Order}, \textbf{Parameter}, and \textbf{Nested Param}, we also consider the following metrics:

\textbf{Format} measures the accuracy of LLM's output format.

\textbf{Tree} measures the pass rate of the entire call tree of LLMs, requiring absolute correctness in the four previous metrics of Selection, Order, Parameter, and Nested Param to be considered a pass.

We provide the format accuracy and Tree pass rate of all evaluated LLMs in Table~\ref{tab:format-tree-acc}.

\subsection{Evaluation Prompt}
\label{Evaluation details: Evaluation Prompt}

The corresponding prompt is presented in Table~\ref{tabs: Evaluation Prompt}.

\subsection{Prompt of removing conjunctions}
\label{Evaluation details: Prompt of removing conjunctions}

The corresponding prompt is presented in Table~\ref{tabs: Prompt of removing conjunctions}.

\subsection{Error Examples}
\label{sec:error-examples}

\subsubsection{Tool Selection}
\label{Error examples: Tool Selection}

The error examples for Tool Selection are shown in \tablename~\ref{tab: Error examples for Tool Selection}.

\subsubsection{Tool Omission}
\label{Error examples: Tool Omission}

The error examples for Tool Omission are shown in \tablename~\ref{tab: Error examples for Tool Omission}.

\subsubsection{Tool Redundancy}
\label{Error examples: Tool Redundancy}

The error examples for Tool Redundancy are shown in \tablename~\ref{tab: Error examples for Tool Redundancy}.

\subsubsection{Hallucinations}
\label{Error examples: Hallucinations}

The error examples for Hallucinations are shown in \tablename~\ref{tab: Error examples for Hallucinations}.

\subsubsection{Generation Format}
\label{Error examples: Generation Format}

The error examples for Generation Format are shown in \tablename~\ref{tab: Error examples for Generation Format}.

\begin{table*}[ht]
    \centering
    \begin{tcolorbox}[title=Tool/Instance Generation, colback=gray!20, colframe=gray!75, rounded corners]
        Please generate several realistic functions with definitions in the given specific domain and generate complete function calling statements in `main()`, shaped like "data0,data1 = func0(*args)" , "data2,data3,data4 = func1(*args)",etc. Do not add any additional python code in `main()`.
The most important thing is that the former function's return value should be latter ones' input argument, in other words, data1, data2, etc. should work as the arguments of subsequent function calls.In this scenario, the subsequent function will use a return value of a preceding function as its parameter, you need to ensure that the inner meaning of this parameter in both functions should be consistent, with identical parameter type, while slight difference in parameter names and descriptions are acceptable.\\
All required parameters must be filled in, by directly writing the actual values in order. Optional parameters can be filled in or not filled in, but remember to add "parameter\_{}name=" before the value of the optional parameter if you are sure to fill in this optional parameter.\\
You need to write the task in natural language as a comment under the function calling statements in `main()` to describe the entire calling process. Please ensure that it includes all the function calling values.\\
Don't use python's intrinsic functions in your generation, such as lists and dictionaries (index access, slicing, adding, deleting, etc.). \\
The generated API only needs to be related to the given domain and does not need to be completely subordinate to the given domain.\\

Here are some examples.\\
\{example\} \\

It's your turn to generate. \\
Domain: \{\}\\
Answer: 
        
    \end{tcolorbox}
    \caption{Prompt of Tool/Instance Generation.}
    \label{tabs:Tool/Instance Generation}
\end{table*}

\begin{table*}[ht]
    \centering
    \begin{tcolorbox}[title=An example for Tool/Instance Generation, colback=gray!20, colframe=gray!75, rounded corners]
        \begin{verbatim}
Domain: Food and Beverage/Food product development
```python
def get_top_selling_wine(location: str, date: str):
    """Retrieve the top-selling wine in a specific location on a specific date.

    required parameters:
        location (str): the geographical location (country, region, city)
        date (str): the date of the sales in 'YYYY-MM-DD' format

    Returns:
        wine_name (str): the name of the top-selling wine
        sales_value (float): the sales value of the wine in dollars
    """
    ...
def get_product_origin(product: str):
    """Retrieve the origin of a specific product.

    required parameters:
        product (str): the name or code of the product

    Returns:
        origin (str): the geographical origin of the product
    """
    ...
def get_export_quantity(location: str, date: str, product: str):
    """Retrieve the export quantity of a specific product 
     from a specific location on a specific date.

    required parameters:
        location (str): the geographical location (country, region, city)
        date (str): the date of the export in 'YYYY-MM-DD' format
        product (str): the name or code of the product

    Returns:
        quantity (int): the export quantity of the product
    """
    ...
def main():
    data0, data1 = get_top_selling_wine("California", "2023-10-08")
    data2 = get_product_origin(data0)
    data3 = get_export_quantity(data2, "2023-10-08", data0)
     # I would like to know which wine sold the best in California on October 8th, 
     2023. Could you provide me with the origin of this wine and inform me about 
     the export data of this wine from the local area on that day?
if __name__ == "__main__":
    main()
```
\end{verbatim}
    \end{tcolorbox}
    \caption{An example for Tool/Instance generation process.}
    \label{tabs:ICL example for Tool/Instance Generation}
\end{table*}

\begin{table*}[ht]
    \centering
    \begin{tcolorbox}[title=Query Normalization, colback=gray!20, colframe=gray!75, rounded corners]

        Below are some APIs, a task, and API call result for this task. You need to check if any values of the API call result that are not present in the task, incorporate those values into the original task without mentioning the specific APIs, rewrite the original task to ensure that it includes all the values,which is very important.If the value is in the format of complex data type (such as a dictionary, list, tuple, etc.), you need to incorporate each item from the value into the task in natural language, rather than directly adding the value in the raw format to the task.
        
The instructions of the rewritten task do not need to be completely consistent with the information in the given API such as the name and despcription, and some uniqueness of the real-life scenario should be retained.

You only need to generate the rewritten task without adding any additional explanations.\\

Here is an example.

\{example\}\\

It's your turn to generate.\\

APIs: 

\{gold apis\}

Task: 

\{task\}

API call result: 

\{call chain with gold apis\}

rewritten task:

    \end{tcolorbox}
    \caption{Prompt of query normalization.}
    \label{tabs: Query Normalization}
\end{table*}

\begin{table*}[ht]
    \centering
    \begin{tcolorbox}[title=Query Refinement, colback=gray!20, colframe=gray!75, rounded corners]

        You are an excellent task rewriter. Next, I will give you a task query, and ask you to rewrite it in high quality. Please follow the tips below.
        
Tips:

1. Please make sure that the actual name of the API(e.g., get\_{}article\_{}headline) involved in the call does not appear in the rewritten task.

2. The API call result represents the golden tool calling process corresponding to the query. Parameter value in the "API\_{}call\_{}" format represents a return value from a previous tool. Refer to the provided APIs information to ensure that the rewritten task is consistent with the golden tool calling process.

3. Please make sure that all regular parameters in the gold tool calling process, which are not in the "API\_{}call\_{}" format, can be extracted or inferred from the rewritten task.

4. Without changing the core meaning and logic of the original task, and while adhering to the call chain, appropriately modify the form of expression, make the task more coherent and aligned with real-world scenarios. \\

Here's an example:

\{example\} \\

It's your turn to generate.\\

APIs:

\{gold apis\}

Task: 

\{original task\}

API call result:

\{call chain with gold apis\}

rewritten task:

    \end{tcolorbox}
    \caption{Prompt of query refinement.}
    \label{tabs: Query Refinement}
\end{table*}

\begin{table*}[ht]
\centering
\begin{tabularx}{0.7\textwidth}{l *{3}{>{\centering\arraybackslash}X}}
\toprule
\multicolumn{1}{l}{\multirow{1}{*}{\textbf{Model}}} & \multicolumn{1}{c}{\textbf{Format}} & \multicolumn{1}{c}{\textbf{Tree}}  \\
\midrule
\rowcolor{red!20} \multicolumn{3}{c}{\textit{Proprietary Models}} \\

GPT-4o      & \textbf{100.0} & 25.9  \\
GPT-4o-mini & 94.4 & 18.3 \\
GPT-3.5     & 99.4 & 13.5  \\
Claude-3.5  & \textbf{100.0} & \textbf{26.8}  \\

\midrule
\rowcolor{green!20} \multicolumn{3}{c}{\textit{Open-Weight Models}} \\
LLaMA3.1-8B   & 98.0 & 6.9 \\
LLaMA3.1-70B  & 99.1 & 9.7  \\
LLaMA3.1-405B & 97.5 &  14.7 \\
Mistral-7B v0.2  & 68.8 & 2.9  \\
Mixtral-8x7B  & 91.3 & 7.4  \\
Mixtral-8x22B & \textbf{99.5} & 17.5  \\
Qwen2-0.5B    & 1.2  & 0.0   \\
Qwen2-1.8B    & 66.5 & 0.0  \\
Qwen2-7B      & 94.2 & 4.6  \\
Qwen2-57B     & 97.5 & 9.2  \\
Qwen2-72B     & 98.6 & \textbf{20.4}  \\
Qwen1.5-0.5B  & 3.2 & 0.0   \\
Qwen1.5-1.8B  & 12.6 & 0.0   \\
Qwen1.5-7B    & 59.4 & 2.4  \\
Qwen1.5-14B   & 85.4 & 7.9  \\
Qwen1.5-32B   & 85.6 & 10.7  \\
Qwen1.5-72B   & 86.1 & 7.8  \\
Qwen1.5-110B  & 88.8 & 13.0 \\
\bottomrule
\end{tabularx}
\caption{Format accuracy and Tree pass rate of \nestools. \textbf{Bold} represents the best score among all models.}
\label{tab:format-tree-acc}
\end{table*}
\label{table: other metrics}

\begin{table*}[ht]
    \centering
    \begin{tcolorbox}[title=Prompt of Evaluation, colback=gray!20, colframe=gray!75, rounded corners]

    You have access to a list of APIs and the task description.You need to follow the given task description and determine which API to call in sequence according to the order required by the task description. API can be retrieved from the APIs list. Finally, you only need to return the API call result without any other content.
    
The final result should be in the format of [\{"api\_{}name":\_{}\_{},"api\_{}id":\_{}\_{},"parameters":\{"arg0":"value0",

"arg1":"value1",...\},"responses":\{"arg0":"API\_{}call\_{}0", ... ,"argn":"API\_{}call\_{}n"\}\},\{"api\_{}name":\_{}\_{},

"api\_{}id":\_{}\_{},"parameters":\{"arg0":"value0","arg1":"value1",...\},"responses":\{"arg0":"API\_{}call\_{}\{n+1\}",

...\}\}, ...].

You don't need to know the actual return value of the API call, just assign each return value as a string "API\_{}call\_{}\{number\}" in "responses", such as "API\_{}call\_{}0","API\_{}call\_{}1","API\_{}call\_{}2" and so on. The "number" in "API\_{}call\_{}\{number\}" should increase by one from 0 globally. 

Please first determine which APIs to call in sequence based on the task, and then determine the parameter values of each API depending on the specific details of the task. If you decide to call the API, you need to fill in all this API's required parameters which can be found in this API's "required" list. If you think the task does not include the actual value of a necessary parameter in API's "required" list, you can assign the necessary parameter a value of "UNK". The remaining parameters are optional parameters, determine whether to fill them in according to the task. If you think the parameter value to be filled in is the return value of a previous API call, set it as "API\_{}call\_{}x", then the parameter value can be filled in with "API\_{}call\_{}x".\\

Now it is your turn to generate the API call result based on the APIs and task description below. Remember that you only need to generate the API call result, not any additional explanations.

APIs:

\{tools\}

Task description:

\{task description\}

API call result:

    \end{tcolorbox}
    \caption{Prompt of evaluation.}
    \label{tabs: Evaluation Prompt}
\end{table*}

\begin{table*}[ht]
    \centering
    \begin{tcolorbox}[title=Remove Conjunctions, colback=gray!20, colframe=gray!75, rounded corners]

        Now there is a task description. I hope you can remove potential conjunctions like "Please", "then", "subsequently", "After that" etc., while maintaining the core meaning. Ensure that the text remains in natural language without punctuation errors and adjust the capitalization accordingly.\\
        
Here is an example:

\{example\}\\

Now its your turn:

Task description: \{\}

Result:

    \end{tcolorbox}
    \caption{Prompt of removing conjunctions.}
    \label{tabs: Prompt of removing conjunctions}
\end{table*}

\begin{table*}
    \centering
    \setlength{\tabcolsep}{3pt}
    \resizebox{\textwidth}{!}{\begin{tabular}{>{\centering\arraybackslash}p{0.16\linewidth}|p{0.4\linewidth}|p{0.5\linewidth}}
        \hline
        \rowcolor{gray!20} 
        \textbf{Model} & \centering\textbf{Query} & \textbf{Response} \\
        \hline
        Qwen2-7B & I currently require a brand new red medium-sized dress. Please assist in designing the layout and selecting a luxurious silk fabric for this dress. Finally, arrange for Taylor Smith to bring the entire design to life and maintain oversight of the progress. & [\{"api\_{}name": "design\_{}garment", "parameters": \{"garment\_{}type": "dress", "color": "red", "size": "medium"\}, "responses": \{\}\}, \{"api\_{}name": "fabric\_{}selection", "parameters": \{"design": "A red medium-sized dress", "fabric\_{}type": "luxurious silk"\}, "responses": \{\}\}, \textcolor{red}{\{"api\_{}name": "choose\_{}manufacturer", "parameters": \{"design\_{}sketch": "A red medium-sized dress made of luxurious silk", "material\_{}list": ["luxurious silk"], "location": "Taylor Smith's workshop"\}, "responses": \{\}\}}]
        \\
        \hline
        LLaMA3.1-8B & Please share the title of a piece in the politics division that incorporates the word which is elections. Following this, extract information concerning the author who penned the aforementioned piece. Lastly, suggest 5 articles that bear similarity about elections. & 
       [\textcolor{red}{\{"api\_{}name": "request\_{}article", "parameters": \{"article\_{}title": "piece in the politics division that incorporates the word which is elections", "author": "UNK", "access\_{}link": "UNK"\}, "responses": \{"access\_{}granted": "API\_{}call\_{}0", "download\_{}link": "API\_{}call\_{}1"\}\}}, \{"api\_{}name": "get\_{}author\_{}info", "parameters": \{"author": "API\_{}call\_{}1"\}, "responses": \{"birthplace": "API\_{}call\_{}2", "articles\_{}written": "API\_{}call\_{}3"\}\}, \{"api\_{}name": "recommend\_{}similar\_{}articles", "parameters": \{"keyword": "elections", "limit": "5"\}, "responses": \{"similar\_{}articles": "API\_{}call\_{}4"\}\}] \\
        \hline
    \end{tabular}}
    \caption{Error examples for tool selection.}
    \label{tab: Error examples for Tool Selection}
\end{table*}

\begin{table*}
    \centering
    \setlength{\tabcolsep}{3pt}
    \resizebox{\textwidth}{!}{\begin{tabular}{>{\centering\arraybackslash}p{0.16\linewidth}|p{0.4\linewidth}|p{0.5\linewidth}}
        \hline
        \rowcolor{gray!20}
        \textbf{Model} & \centering\textbf{Query} & \textbf{Response} \\
        \hline
        LLaMA3.1-70B & Scan the ISBN "978-3-16-148410-0" of a book to extract comprehensive data and verify its presence in the library. Upon identifying the book situated, launch an augmented reality interaction correlating to the book. &  
        [\{"api\_{}name": "scan\_{}isbn", "parameters": \{"isbn": "978-3-16-148410-0"\}, "responses": \{"book\_{}details": "API\_{}call\_{}0", "availability": "API\_{}call\_{}1"\}\}, \textcolor{red}{\sout{\{"api\_{}name": "locate\_{}book", "parameters": \{...\}, "responses": [...]\}}},\{"api\_{}name": "engage\_{}ar\_{}experience", "parameters": \{"availability": "API\_{}call\_{}1", "exact\_{}location": "API\_{}call\_{}0"\}, "responses": \{"ar\_{}message": "API\_{}call\_{}2", "ar\_{}duration": "API\_{}call\_{}3"\}\}]
        \\
        \hline
        Qwen2-7B & I would like to know which company is responsible for producing the Smartphone X10, and I am also interested in understanding the comprehensive development value of the industrial zone it belongs to. & 
        [\{"api\_{}name": "get\_{}product\_{}manufacturer", "parameters": \{"product\_{}name": "Smartphone X10"\}, "responses": \{"manufacturer\_{}name": "API\_{}call\_{}0"\}\}, \textcolor{red}{\sout{\{"api\_{}name": "get\_{}factory\_{}location", "parameters": \{...\}, "responses": [...]\}}},\{"api\_{}name": "evaluate\_{}property", "parameters": \{"location": "API\_{}call\_{}0", "type\_{}of\_{}land": "industrial", "area": 100.0\}, "responses": \{"land\_{}value": "API\_{}call\_{}1", "development\_{}potential": "API\_{}call\_{}2"\}\}] \\
        \hline
    \end{tabular}}
    \caption{Error examples for tool omission.}
    \label{tab: Error examples for Tool Omission}
\end{table*}

\begin{table*}
    \centering
    \setlength{\tabcolsep}{3pt}
    \resizebox{\textwidth}{!}{\begin{tabular}{>{\centering\arraybackslash}p{0.16\linewidth}|p{0.4\linewidth}|p{0.5\linewidth}}
        \hline
        \rowcolor{gray!20}
        \textbf{Model} & \centering\textbf{Query} & \textbf{Response} \\
        \hline
        LLaMA3.1-8B & I'm planning a metro communication route that needs to span about 50 km, and we're targeting a bandwidth capacity of 100 Gbps. Could you please provide a modulation format that will work well over that distance. Also, there's the challenge of compensating for a line loss of around 3.5 dB, so I'll need to fine-tune the settings accordingly. Once got the right settings, Could you tell me how to set up a stable connection between A1 and B1? & 
        [\{"api\_{}name": "choose\_{}modulation\_{}format", "parameters": \{"route\_{}type": "metro", "distance": 50, "bandwidth": 100\}, "responses": \{"modulation\_{}scheme": "API\_{}call\_{}0", "system\_{}settings": "API\_{}call\_{}1"\}\}, \{"api\_{}name": "adjust\_{}amplification", "parameters": \{"format": "API\_{}call\_{}0", "line\_{}loss": 3.5\}, "responses": \{"optimized\_{}gain": "API\_{}call\_{}2", "amplifier": "API\_{}call\_{}3", "signal\_{}quality": "API\_{}call\_{}4"\}\}, \textcolor{red}{\{"api\_{}name": "setup\_{}receiver", "parameters": \{"trans\_{}config": "API\_{}call\_{}3", "link\_{}budget": "API\_{}call\_{}2", "channel\_{}type": "single-mode"\}, "responses": \{"receiver\_{}status": "API\_{}call\_{}5", "signal\_{}quality": "API\_{}call\_{}6"\}\}}, \{"api\_{}name": "establish\_{}connection", "parameters": \{"amplifier\_{}type": "API\_{}call\_{}3", "end\_{}points": "A1-B1"\}, "responses": \{"connection\_{}status": "API\_{}call\_{}7", "connection\_{}details": "API\_{}call\_{}8"\}\}]
        \\
        \hline
        Qwen2-7B & I'm working on a project called "Alien Planet", which is a VFX scene I aim to create. It should last about 120 seconds and I want it to be in 4K resolution to ensure high quality. I'd like to include SFX featuring alien creatures and spacecraft in a mysterious atmosphere. Once everything is put together, I need the scene to be in MP4 format. Could you guide me with the whole process? & 
        [\{"api\_{}name": "create\_{}vfx\_{}scene", "parameters": \{"scene\_{}name": "Alien Planet", "duration": 120, "resolution": "4K"\}, "responses": \{\}\}, \{"api\_{}name": "add\_{}special\_{}effects", "parameters": \{"scene\_{}path": "path\_{}to\_{}Alien\_{}Planet\_{}scene", "effects\_{}list": ["alien\_{}creatures", "spacecraft"], "mood": "mysterious"\}, "responses": \{\}\}, \textcolor{red}{\{"api\_{}name": "render\_{}scene", "parameters": \{"scene\_{}objects": ["alien\_{}creatures", "spacecraft"], "quality": "high", "resolution": [3840, 2160]\}, "responses": \{\}\}}, \{"api\_{}name": "export\_{}final\_{}scene", "parameters": \{"scene\_{}path": "path\_{}to\_{}rendered\_{}Alien\_{}Planet\_{}scene", "format\_{}type": "MOV"\}, "responses": \{\}\}] \\
        \hline
    \end{tabular}}
    \caption{Error examples for tool redundancy.}
    \label{tab: Error examples for Tool Redundancy}
\end{table*}

\begin{table*}
    \centering
    \setlength{\tabcolsep}{3pt}
    \resizebox{\textwidth}{!}{\begin{tabular}{>{\centering\arraybackslash}p{0.16\linewidth}|p{0.4\linewidth}|p{0.5\linewidth}}
        \hline
        \rowcolor{gray!20}
        \textbf{Model} & \centering\textbf{Query} & \textbf{Response} \\
        \hline
        LLaMA3.1-8B & Please assist in designing an optimal prosthesis for a 35-year-old male, who has an unremarkable medical past and has suffered a fracture impacting his upper extremity. Afterward, evaluate the patient's steady condition and anticipate a complete recovery within a timeframe of 6 months. Finally, arrange an orthopedic surgical procedure for the patient. & 
        [\{"api\_{}name": "design\_{}prosthesis", "parameters": \{"patient\_{}info": "\{"age": 35, "gender": "male", "medical\_{}history": "unremarkable"\}", "injury": "fracture", "affected\_{}area": "upper extremity"\}, "responses": \{"prosthesis\_{}details": "API\_{}call\_{}0", "estimated\_{}cost": "API\_{}call\_{}1"\}\}, \{"api\_{}name": "patient\_{}assessment", "parameters": \{"patient\_{}condition": "stable", "recovery\_{}expectations": "complete recovery within 6 months"\}, "responses": \{"assessment\_{}report": "API\_{}call\_{}2"\}\}, \textcolor{red}{\{"api\_{}name": "schedule\_{}orthopedic\_{}surgery", "parameters": \{"prosthesis": "API\_{}call\_{}0", "patient\_{}info": "\{\"age\": 35, \"gender\": \"male\", \"medical\_{}history\": \"unremarkable\"\}", "surgery\_{}type": "orthopedic", "surgery\_{}date": "2024-03-16"\}, "responses": \{"surgery\_{}schedule": "API\_{}call\_{}3"\}\}}]
        \\
        \hline
        Qwen2-7B & Assess the caliber of the Python code in the file found at "path/to/file.py". Subsequently, refine the code using the tool PyLint. Finally, deploy the optimized code as a web application to the production environment. & 
        [\{"api\_{}name": "analyze\_{}code\_{}quality", "parameters": \{"file\_{}path": "path/to/file.py", "language": "Python"\}, "responses": \{"issues\_{}found": ["issue1", "issue2"], "complexity\_{}score": 4.5, "maintainability\_{}score": 8.2\}\}, \textcolor{red}{\{"api\_{}name": "optimize\_{}code", "parameters": \{"issues": ["issue1", "issue2"], "optimization\_{}tool": "PyLint"\}, "responses": \{"optimized\_{}code": "optimized\_{}code\_{}snippet"\}\}}, \{"api\_{}name": "deploy\_{}app", "parameters": \{"app\_{}id": "optimized\_{}code\_{}snippet", "server\_{}type": "Production", "region": "Global"\}, "responses": \{"deployment\_{}status": true, "deployed\_{}url": "https://optimized-code.example.com"\}\}] \\
        \hline
    \end{tabular}}
    \caption{Error examples for hallucinations.}
    \label{tab: Error examples for Hallucinations}
\end{table*}

\begin{table*}
    \centering
    \setlength{\tabcolsep}{3pt}
    \resizebox{\textwidth}{!}{\begin{tabular}{>{\centering\arraybackslash}p{0.16\linewidth}|p{0.4\linewidth}|p{0.5\linewidth}}
        \hline
        \rowcolor{gray!20}
        \textbf{Model} & \centering\textbf{Query} & \textbf{Response} \\
        \hline
        LLaMA3.1-405B & Please check the safety status of a car with registration number ABC123, last inspected on May 12, 2022. Issue a class C driving license to John Doe aged 30. Then book a vehicle inspection appointment for a truck with registration number DEF456 on June 20, 2022. & 
        [\{"api\_{}name": "check\_{}vehicle\_{}safety", "parameters": \{"vehicle\_{}type": "car", "registration\_{}number": "ABC123", "inspection\_{}date": "May 12, 2022"\}, "responses": \{"safety\_{}status": "API\_{}call\_{}0", "next\_{}inspection\_{}date": "API\_{}call\_{}1"\}\}, \textcolor{red}{\{"api\_{}name": "issue\_{}driving\_{}license", "parameters": \{"applicant\_{}name": "John Doe", "age": 30, "license\_{}type": "class C"\}\}}, "responses": \{"license\_{}number": "API\_{}call\_{}2", "expiration\_{}date": "API\_{}call\_{}3"\}\}, \{"api\_{}name": "book\_{}vehicle\_{}inspection", "parameters": \{"registration\_{}num": "DEF456", "vehicle": "truck", "inspection\_{}date": "June 20, 2022"\}, "responses": \{"appointment\_{}confirmation": "API\_{}call\_{}4"\}\}]
        \\
        \hline
        Qwen2-7B & Load the VR environment named "Fantasy Land" for user ID 12345. Obtain information about the user's avatar named "Wizard". Explore the VR world within the "Fantasy Land" environment, starting at coordinates (10, 5, 3) and interact with other users present. & 
        [\{"api\_{}name": "load\_{}vr\_{}environment", "parameters": \{"environment\_{}name": "Fantasy Land", "user\_{}id": 12345\}, "responses": \{"environment\_{}type": "Fantasy", "user\_{}preferences": "adventure", "avatar\_{}skin\_{}color": "blue"\}\}, \textcolor{red}{\{"api\_{}name": "get\_{}avatar\_{}info", "parameters": \{"user\_{}ID": 12345, "avatar\_{}name": "Wizard"\}, "responses": \{"avatar\_{}height": 1.8, "avatar\_{}accessories": "pointed hat, staff", "avatar\_{}bio": "A powerful wizard who specializes in elemental magic."\}}, \{"api\_{}name": "explore\_{}vr\_{}world", "parameters": \{"environment\_{}type": "Fantasy", "user\_{}coordinates": [10, 5, 3], "interact\_{}with\_{}others": true\}, "responses": \{"discovered\_{}elements": ["ancient ruins", "enchanted forest", "dragon's lair"], "interactive\_{}users": 5\}\}] \\
        \hline
    \end{tabular}}
    \caption{Error examples for generation format.}
    \label{tab: Error examples for Generation Format}
\end{table*}

\end{document}